\documentclass{article}
\usepackage{multirow}

\usepackage{PRIMEarxiv}

\usepackage[utf8]{inputenc} 
\usepackage[T1]{fontenc}    
\usepackage{hyperref}       
\usepackage{url}            
\usepackage{booktabs}       
\usepackage{amsfonts}       
\usepackage{nicefrac}       
\usepackage{microtype}      
\usepackage{lipsum}
\usepackage{fancyhdr}       
\usepackage{graphicx}       
\graphicspath{{media/}}     
\usepackage{float}
\usepackage{amsmath}
\usepackage{orcidlink}
\usepackage{hyperref}

\title{Colors Matter: AI-Driven Exploration of Human Feature Colors
\thanks{\textit{\underline{Citation}}: 
\textbf{R. Alyoubi ,T. Alharbi ,A. Alghamdi, Y. Alshehri and E. Alghamdi Colors Matter: AI-Driven Exploration of Human Feature Colors. To appear.}} 
}

\author{
\begin{tabular}{c}
\textbf{Rama Alyoubi}~\orcidlink{0009-0009-8445-0511} \\
\texttt{Rama.mohammed.alyoubi@gmail.com}
\end{tabular}
\hspace{2cm}
\begin{tabular}{c}
\textbf{Taif Alharbi}~\orcidlink{0009-0004-7044-3656} \\
\texttt{AI.Taif.Alharbi@gmail.com}
\end{tabular}
\\[0.5cm]
\begin{tabular}{c}
\textbf{Albatul Alghamdi}~\orcidlink{0009-0003-0305-6448} \\
\texttt{Albatulalghamdi1@gmail.com}
\end{tabular}
\hspace{2cm}
\begin{tabular}{c}
\textbf{Yara Alshehri}~\orcidlink{0009-0002-2440-3624} \\
\texttt{Yara55hassan@gmail.com}
\end{tabular}
\\[0.5cm]
\begin{tabular}{c}
\textbf{Elham Alghamdi} \\
\texttt{ealghamdi2@uj.edu.sa}
\end{tabular}
}

\begin{document}
\maketitle

\begin{abstract}
This study presents a robust framework that leverages advanced imaging techniques and machine learning for feature extraction and classification of key human attributes—namely skin tone, hair color, iris color, and vein-based undertones. The system employs a multi-stage pipeline involving face detection, region segmentation, and dominant color extraction to isolate and analyze these features. Techniques such as X-means clustering, alongside perceptually uniform distance metrics like Delta E (CIEDE2000), are applied within both LAB and HSV color spaces to enhance the accuracy of color differentiation. For classification, the dominant tones of the skin, hair, and iris are extracted and matched to a custom tone scale, while vein analysis from wrist images enables undertone classification into "Warm" or "Cool" based on LAB differences. Each module uses targeted segmentation and color space transformations to ensure perceptual precision. The system achieves up to \textbf{80\% }accuracy in tone classification using the Delta E–HSV method with Gaussian blur, demonstrating reliable performance across varied lighting and image conditions. This work highlights the potential of AI-powered color analysis and feature extraction for delivering inclusive, precise, and nuanced classification, supporting applications in beauty technology, digital personalization, and visual analytics.
\end{abstract}

\keywords{Skin Tone \and Deep Learning \and Human Skin Detection \and Segmentation \and Color Spaces \and Machine Learning \and Color Distance \and Image Processing \and CIEDE2000 \and Skin Tone Dataset}

\section{Introduction}
Human beauty represents a harmonious combination of skin tone, hair color, and eye color, with each nuanced shade telling a deeper story of personal identity, culture, and individuality. In this context, color transcends its basic visual function, serving as a significant indicator of uniqueness. However, the accurate identification and categorization of these natural tones present a substantial challenge. Unlike the straightforward task of recognizing the color of inanimate objects, human features display subtle variations influenced by factors such as lighting, environment, and genetics. This complexity necessitates more than simple color detection; it requires sophisticated systems capable of interpreting nuanced context and variation. In this study, we introduce an innovative framework that integrates computer vision and machine learning to classify human skin, hair, and eye colors with high accuracy. Through careful preprocessing and the assessment of various models, our approach fosters inclusive and intelligent solutions relevant to personalized beauty, digital representation, and beyond.

\section{Literature Review}
\subsection{Background}

\subsubsection{Skin Undertone and Overtone}
Skin overtone refers to the visible color of the skin, determined by melanin levels in the top layer, ranging from light to dark. In contrast, skin undertone is the subtle hue beneath the surface, which remains consistent despite external changes to skin tone. Undertones are categorized as warm (yellow, peach, or gold hues), cool (blue, pink, or purple hues), or neutral (a balance of both cool and warm tones). Identifying your undertone is key for choosing complementary makeup shades, clothing colors, and metals\cite{Jackson1980}.

\subsubsection{Overview of Skin Tone Classification Scales}
\begin{figure}[h!]
    \centering
    \includegraphics[width=0.55\textwidth]{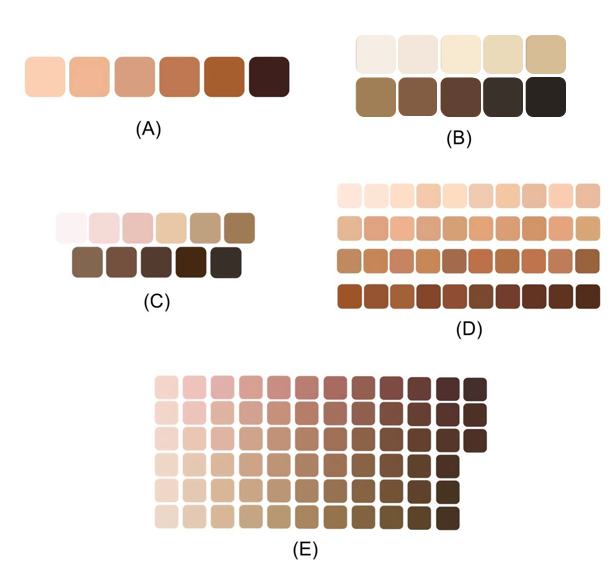}
    \caption{(A) Fitzpatrick Scale. (B) Monk Scale. (C) PERLA Scale. (D) Rihanna Fenty Beauty Scale. (E) Graphic-Based Grid (69-Point) Scale.}
    \label{fig:Scales}
\end{figure}

Skin color representation has traditionally been anchored by systems designed to categorize and define skin tones for a range of applications, from medical treatments to beauty products. One such system is the Fitzpatrick scale, depicted in figure \ref{fig:Scales}. This 6-point graphics-based system is primarily utilized to assess how different skin types react to phototherapy, with a significant focus on white populations. However, it does not adequately capture the diversity of darker skin tones. To address these shortcomings, more inclusive skin tone scales have been developed. For instance, in figure \ref{fig:Scales}, the 10-point Monk scale aims to better represent ethno-racial diversity, and in figure \ref{fig:Scales} 40-point skin tone palette, inspired by a well-known makeup line, offers a wide array of colors and undertones tailored for darker skin tones. This palette provides four times as many options as the Monk scale, significantly improving the representation of diverse skin tones\cite{heldreth2023skin}. Additionally, in the social sciences, the Text-Based 5-point Scale categorizes skin tone from very light to very dark but often falls short of capturing full diversity. To enhance this, the Graphic-Based 10-Point Scale by Massey and Martin and in figure \ref{fig:Scales} the Graphic-Based Grid (69 Point) Scale, adapted from L'Oréal’s extensive skin tone grid, have been introduced. The latter includes additional shades to better represent darker skin tones, offering a comprehensive and inclusive range of skin tones across the spectrum \cite{campbell2020picture}. Moreover, in figure \ref{fig:Scales}, the 11-point PERLA skin tone scale has been incorporated into various studies, including the New Immigrant Survey, the 2012 General Social Survey, the AmericasBarometer, and the Project on Ethnicity and Race in Latin America \cite{telles2014pigmentocracies}.

\newpage

\subsubsection{Understanding Color Spaces}
Color spaces allow for framing colors, which is very important in both digital imaging and color analysis. The RGB color space, developed based on red, green, and blue channels, best suits electronic displays; hence, for digital applications, it is a standard choice \cite{ibraheem2012understanding}. On the other hand, the CIE-XYZ color space, developed by the International Commission on Illumination, has an excellent correlation with human color perception and serves as the basis for color measurement and calibration \cite{ibraheem2012understanding,chen2016skin}. The YCbCr color space separates the brightness information from color information and therefore is helpful in many image processing and color editing applications by independently controlling luminance and chrominance \cite{chen2016skin,bunte2011learning}. HSV, organizing colors by hue, saturation, and value, is intuitive for color picking and, thus, useful for skin detection tasks that can identify skin tone effectively \cite{chen2016skin,kolkur2017human}. Finally, the CIE-Lab and CIE-Luv color models represent color in a perceptually uniform manner, which can be used for applications that require accurate color matching and analysis, such as the personal color recommendations developed \cite{mamat2018silhouette}.

\subsubsection{Challenges and Innovations in Skin Detection and Undertone Classification}
Skin detection differentiates skin from non-skin pixels in images or videos, prioritizing computational efficiency and resistance to rotation, scaling, and occlusion. Despite its popularity, skin detection encounters challenges, including varying illumination, ethnic diversity, background interference, makeup, hairstyles, eyewear, shadow effects, and movement, all of which complicate accurate pixel identification \cite{Zuo2017,Zhu2022}. Infrared and spectral imaging systems provide partial solutions, yet their high costs and complexity hinder widespread use \cite{inbook}. The selection of color space and skin color model is crucial for effective skin detection, with recent approaches exploring diverse color spaces and techniques for feature extraction and classification. However, substantial color overlap between skin and non-skin pixels, along with similarly colored background objects, complicates differentiation without contextual analysis \cite{Zuo2017}.
Computer vision in fashion and cosmetology introduces promising applications, particularly in vein color analysis for undertone classification. This involves photographing the wrist area to classify vein color as warm, cool, or neutral, enhancing personalized style and makeup choices. The integration of computer vision and machine learning enables precise, individualized beauty recommendations, reflecting the potential of technology to transform personal aesthetic decisions.

\subsubsection{Advancements in Skin Detection Technology}
Skin detection methods have improved significantly, and algorithms have increased accuracy across a wide range of lighting and demographic conditions. Traditional techniques comprising the HSV color model coupled with logical filters have shown merit under a wide range of diverse scenarios; however, such methods have yet to be tested on a broader set of datasets to assure reliability \cite{hassan2023hsv}. More recently, the combination of convolutional and recurrent neural networks has been applied to skin detection to leverage local and contextual information; testing its efficiency in real-world applications remains challenging \cite{Zuo2017}.\\
Generative Adversarial Networks (GANs) have been employed in the creation of more robust datasets that allow for better generalization by models across skin tones, but they have scalability challenges \cite{dash2020biometric}. Another more recent novel approach has involved color and texture analysis, which seems to do better in skin recognition tasks than color-only analysis, besides proving helpful for applications that range from personalized color recommendations down to content moderation \cite{nidhal2009skin}. These techniques make it possible to further adapt skin detection to varying illumination through the use of dynamic skin detection techniques, proposed in more recent unsupervised machine learning studies and are an essential ingredient when considering real-world applications related to personalized beauty and cosmetics \cite{ ISLAM2024104046}.\\
The different research studies conducted on skin color analysis using LAB color space introduce a perceptually uniform approach by minimizing lighting effects, which shows better skin tone mapping relevant to makeup and skincare applications \cite{ srt70088}. These are changes within the line of inclusive, adaptive, and practical skin detection approaches. \\
Yan et al.'s recent work \cite{ KIM2023108247} on representative skin regions is promising, using the dataset of Humanae to find the most color-representative facial regions, such as the central cheeks and below the lips. These insights allow for exact color matching, which is very important for personalized makeup recommendations. Dynamic clustering further has the advantage of adaptively responding to changes in lighting during skin detection, thus enabling interactive applications for beauty and cosmetics.

\subsection{Skin Tone Classification Techniques}       Recent advances in skin tone classification utilize machine learning and hyperspectral imaging to improve accuracy, fairness, and the customization of medical treatments, blending technology with insights from social sciences and healthcare. The Classification Algorithm for Skin Color (CASCo) \cite{sobhan2022subject} is a significant development, employing face detection and k-means clustering for objective skin tone analysis, surpassing traditional limitations. Hyperspectral imaging, combined with machine learning, enables precise skin color classification \cite{gulati2023beautifai}, paving the way for personalized medical and cosmetic applications. Survey research highlights the need for inclusive scales, such as the Fenty Beauty palette and the Monk scale, to reduce AI bias and promote inclusivity \cite{Weiser}. These metrics are essential to representing human skin diversity in technology. Furthermore, deep learning in healthcare, such as the SmartPhone Oxygenation Tool (SPOT) for wound imaging, takes melanin’s impact into account to improve personalized wound care \cite{article2}.

\subsection{Distance Metric in Color Analysis}
Color distance, also known as color difference, is a quantitative measure of the perceptual distinction between two colors in a given color space. This measure is fundamental in fields such as computer vision, printing, and product design, where precise color matching or differentiation is essential. Traditional approaches often rely on simple metrics, but advancements in the field have introduced more perceptually uniform formulas to better align with human visual perception.

\begin{enumerate}
\item \textbf{Euclidean Distance:}
\\Euclidean distance is one of the simplest and most commonly used methods for measuring color differences. In a color space such as RGB or CIELAB, the distance between two colors is computed as:

\[
d = \sqrt{(x_2 - x_1)^2 + (y_2 - y_1)^2 + (z_2 - z_1)^2}
\]

While easy to compute, Euclidean distance has notable limitations. It assumes equal weight for all dimensions, neglecting the non-uniform perceptual sensitivities inherent in most color spaces. As a result, it may fail to accurately reflect human perception.

\item \textbf{CIE Color Difference Metrics:}
\\To address the limitations of Euclidean distance, the Commission Internationale de l'Éclairage (CIE) developed advanced metrics designed to align more closely with perceptual uniformity:

\begin{itemize}
    \item \textbf{CIELAB (\( \Delta E^*_{ab} \))}: Measures color differences in the CIELAB color space, assuming uniform perceptual response. However, it often struggles with highly saturated or near-neutral colors.
    \item \textbf{CIE94}: An improvement over CIELAB, it introduces weighting factors for lightness, chroma, and hue, offering a more accurate perceptual representation.
    \item \textbf{CIEDE2000 (\( \Delta E_{00} \))}: A more advanced metric that builds on \( \Delta E^*_{ab} \) and CIE94 by addressing perceptual non-uniformities, particularly in chroma and hue.
\end{itemize}

\item \textbf{CIEDE2000 and Its Advantages:}
\\The \textbf{CIEDE2000} formula, introduced in 2001, has become widely adopted for its improved perceptual accuracy. It incorporates parametric adjustments and accounts for:

\begin{itemize}
    \item \textbf{Chroma weighting}: Adjusts for the varying impact of chroma differences on perception.
    \item \textbf{Hue weighting}: Captures nonlinearities in hue perception.
    \item \textbf{Interaction terms}: Addresses the interplay between chroma and hue.
\end{itemize}

The formula is expressed as:

\[
\Delta E_{00} = \sqrt{\left( \frac{\Delta L'}{k_L S_L} \right)^2 + \left( \frac{\Delta C'}{k_C S_C} \right)^2 + \left( \frac{\Delta H'}{k_H S_H} \right)^2 + R_T \left( \frac{\Delta C'}{k_C S_C} \right) \left( \frac{\Delta H'}{k_H S_H} \right)}
\]

Where:
\begin{itemize}
    \item \( \Delta L' \): Lightness difference,
    \item \( \Delta C' \): Chroma difference,
    \item \( \Delta H' \): Hue difference,
    \item \( S_L, S_C, S_H \): Scaling factors for lightness, chroma, and hue,
    \item \( R_T \): A rotation term accounting for chroma-hue interaction.
\end{itemize}

\item \textbf{Application of CIEDE2000:}
\\The \textbf{CIEDE2000} metric was utilized to quantify color differences due to its precision and alignment with perceptual relevance. This metric proved particularly effective for tasks such as color conversion and matching, ensuring results closely mirrored human visual assessment. Its robustness and adaptability make it an ideal choice for modern applications requiring high fidelity in color analysis.

Furthermore, the CIEDE2000 formula resolves discontinuities and computational errors inherent in earlier metrics\cite{sharma2005ciede2000}. These features solidify its position as a superior tool for consistent and reliable color analysis.
\end{enumerate}

\section{Dataset}
\subsection{Skin Tone Dataset}
\label{sec:Dataset/SkinToneDataset}
To develop and evaluate the performance of our classification system, we compiled a diverse set of facial images representing a wide range of skin tones. The dataset was constructed to ensure balance, inclusivity, and variability in lighting and image conditions, which are essential for training robust models. All datasets used are available in the \href{https://github.com/AiTaif7/Color-Matters-AI-Driven-Exploration-of-Human-Feature-Coloration/tree/main}{GitHub repository}.

The dataset is divided into two sets:
\begin{itemize}
    \item \textbf{First Dataset}: 720 images categorized into 8 classes according to our new skin tone scale. Table \ref{tab:dataset1}.
    \item \textbf{Second Dataset}: 400 images, evenly distributed across the same 8 classes.Table \ref{tab:dataset2}.
\end{itemize}

\begin{table}[h!]
\centering
\renewcommand{\arraystretch}{1.2} 
\begin{tabular}{p{3cm} p{4cm} p{1.8cm} p{3.5cm}}
\toprule
\textbf{Source} & \textbf{Description} & \textbf{Total Images} & \textbf{Class Distribution} \\
\midrule
Face Research Lab London & Real people, controlled lighting and poses & 88 & 
\begin{tabular}[c]{@{}l@{}} 
Class 4: 19 images \\ 
Class 6: 37 images \\ 
Class 8: 32 images 
\end{tabular} \\
\addlinespace
\addlinespace
Flickr Faces HQ (FFHQ) & Real people, varied lighting and poses & 312 & 
\begin{tabular}[c]{@{}l@{}} \\
Classes 1,2,3,5,7:50 \\ images each \\ 
Class 4: 31 images \\ 
Class 6: 13 images \\ 
Class 8: 18 images 
\end{tabular} \\
\addlinespace
\addlinespace
Synthetic Faces High Quality (SFHQ) & AI-generated faces & 320 & 
Classes 1--8: 40 images each \\
\bottomrule
\end{tabular}
\caption{Summary of Dataset 1 sources and class distributions.}
\label{tab:dataset1}
\end{table}

\begin{table}[h!]
\centering
\renewcommand{\arraystretch}{1.2} 
\begin{tabular}{p{3.5cm} p{5.2cm} p{3cm}}
\toprule
\textbf{Source} & \textbf{Description} & \textbf{Class Distribution} \\
\midrule
Flickr Faces HQ (FFHQ) & Real people, varied lighting and poses & Classes 1--8: 50 images each \\
\bottomrule
\end{tabular}
\caption{Summary of Dataset 2 sourced from FFHQ.}
\label{tab:dataset2}
\end{table}

\noindent In total, the two datasets encompass a broad diversity of skin tones under various lighting conditions, supporting the development of robust classification models.

\newpage

\subsection{Iris, Hair and Undertone Dataset}
For the purpose of preliminary code testing, approximately 10 images were manually collected for each color category within the features of iris, hair, and veins from publicly available sources on the internetr. The images represent different color variations and were used solely to verify the initial functionality of the implemented algorithms. These images were not part of any structured data set and were not used for model training or final evaluation.

\section{Methodology}
\begin{figure}[h!]
\centering
\includegraphics[width=0.9\textwidth]{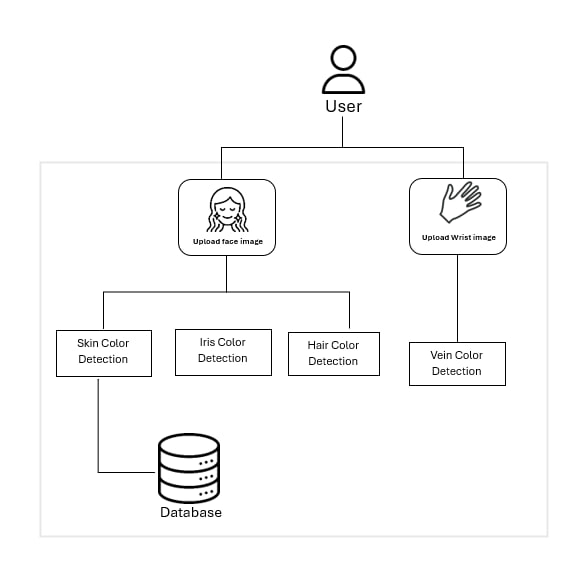}
\caption{System Architecture and Design}
\label{fig:System design}
\end{figure}

The system follows a structured process that begins when users upload a face image and a wrist image through an interactive interface. These images are analyzed to identify key visual features such as skin tone, hair color, eye color, and undertone. Specifically, the face image is processed to detect the dominant skin tone, hair shade, and iris color by focusing on particular regions and analyzing their color patterns. Meanwhile, the wrist image is used to determine the undertone by identifying the color of visible veins. To ensure accurate detection, even under varying lighting conditions, the system employs color comparison techniques and image enhancement methods. All results are then transmitted to a central server for data storage and further use, forming the foundation for personalized, color-based analysis. The implementation of this process can be explored in detail in the \href{https://github.com/AiTaif7/Color-Matters-AI-Driven-Exploration-of-Human-Feature-Coloration/tree/main}{GitHub repository}. Figure \ref{fig:System design}.

\newpage

\subsection{Skin Tone Classification}
\begin{figure}[h!]
\centering
\includegraphics[width=0.9\textwidth]{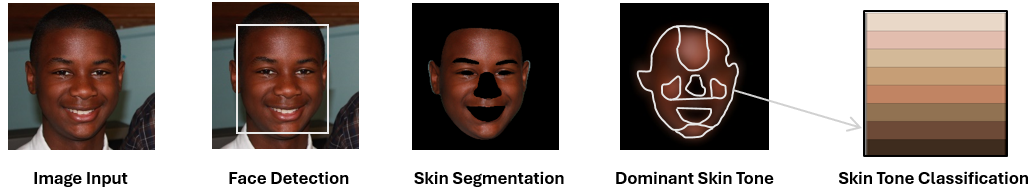}
\caption{Skin Tone Classification Processes, adapted from \cite{Karras2018}}
\label{fig: Skin Tone Classification Processes.}
\end{figure}

In our system, the classification of skin tone plays a crucial role. The process involves a sophisticated multi-step approach, ensuring precise identification and categorization of skin tones from digital images. This section outlines the multi-step methodology for skin tone classification, demonstrated in the uploaded image. The process is shown in Figure \ref{fig: Skin Tone Classification Processes.}.
\begin{enumerate} 

\item \textbf{Image Input:} The system captures and uploads the image, standardizing its size and format for consistent processing.
\item \textbf{Face Detection:} The algorithm detects faces using the Facer library, isolating facial regions and annotating them for model training in skin tone analysis.
\item \textbf{Skin Detection and Segmentation:} Skin areas are isolated by filtering out non-skin elements, ensuring pure skin samples for accurate tone classification.
\item \textbf{Find Dominant Skin Tone:} The main skin tone is identified by applying X-means clustering on HSV values, capturing the region’s primary color.
\item \textbf{Skin Tone Classification:} The dominant tone is classified using our custom skin tone scale, converting it to LAB and finding the closest match based on CIEDE2000 values.
\end{enumerate}

To achieve this, our system employs a series of advanced technological processes, each designed to enhance the accuracy and efficiency of skin tone classification. Let’s delve into the specific technologies and techniques utilized at each stage of the process.\\
\textbf{First Step: Image Input}
This is the initial step in which the system takes a picture and enters it. File uploads can be used to obtain images. The significance of image quality and resolution lies in its direct influence on the efficacy of later stages of processing. At this point, the system incorporates preprocessing processes to standardize image sizes and formats, which aid in preserving uniformity throughout the dataset for precise processing.\\
\textbf{Second Step: Face Detection}
At this stage, the algorithm identifies human faces within the images, isolating skin regions from the eyes, nose, and other background features. This step is crucial for systems designed to analyze characteristics unique to human subjects, such as skin tone classification.
For advanced face detection and parsing techniques, we employ the Facer library, which uses a deep learning model to detect facial features. Facer is essential for isolating facial regions and handling complex backgrounds, making it a robust solution for face detection across varying lighting conditions and poses, a critical factor in our diverse dataset \cite{deng2020retinaface}.\\
\textbf{Third Step: Skin Detection and Segmentation}
After detecting faces, the system isolates skin regions by filtering out non-skin elements. We employ Timm for facial parsing and feature extraction. This library offers a collection of pre-trained models that handle different facial attributes with minimal preprocessing, maintaining high throughput without sacrificing data quality \cite{zheng2022general}.\\
\textbf{Fourth Step: Find Dominant Skin Tone}
The dominant skin tone is extracted from isolated blurred skin regions using X-means clustering on HSV values. This process aims to identify the most prominent color in the region, which serves as the primary color representing the skin tone in a simplified format. The following steps outline this process in detail:
\begin{enumerate}

\item \textbf{Cluster Initialization:} The process begins by initializing clusters using the k-means++ method with an initial cluster count set to 2. This initialization provides starting points that help the X-means clustering algorithm achieve better accuracy in identifying distinct color groups within the skin region.
\item \textbf{Clustering Process:} Using the initialized clusters, the function applies X-means clustering on the HSV values of the skin region. This clustering approach dynamically determines the optimal number of clusters based on the data distribution, allowing the system to adapt to variations in skin tones within the isolated region. The clustering uses an Euclidean distance metric, which measures the color differences between HSV values to group similar colors effectively.
\item \textbf{Identifying the Dominant Color:} After processing, the method identifies the clusters and their centers, representing the color distributions in the skin region. The largest cluster, which contains the most pixels, is selected as it reflects the most prominent skin tone in the region. The center of this largest cluster is taken as the dominant color, providing a single representative tone for classification.
\item \textbf{Output:} The dominant color is then returned as an integer array, representing the HSV values of the identified skin tone. This dominant tone is essential for the final classification, as it simplifies the skin tone data to a primary color that aligns closely with the actual skin tone of the individual.
This method ensures a reliable and accurate extraction of the dominant skin tone, which is crucial for the subsequent classification stage.
\end{enumerate}
\textbf{Fifth Step: Skin Tone Classification}
After obtaining the dominant skin tone, the method identifies the closest match within the custom skin tone scale, depicted in Figure \ref{fig: Custom Skin Tone Scale} below. This custom scale provides a precise and inclusive range of skin tone categories, enhancing classification accuracy across a diverse population. It is better suited for classification tasks than other scales, such as the Monk Scale (MST) \cite{telles2014pigmentocracies}, which poses challenges for classifiers, particularly with lighter tones, and the Fitzpatrick Scale (FST) \cite{ware2020racial}, which lacks the diversity needed for accurate representation. Even though the tones in the custom scale are gradual, the colors are distinct enough in tone and brightness that a model can find decision boundaries between them, improving classification robustness. The custom scale was developed through a process of trial and error, without being based on any formal study, laboratory research, or professional calibration, but rather through practical experimentation aimed at optimizing model performance across diverse tones.

\begin{figure}[h!]
\centering
\includegraphics[width=0.9\textwidth]{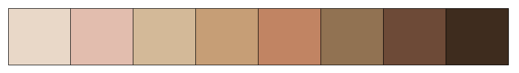}
\caption{Our Custom Skin Tone Scale}
\label{fig: Custom Skin Tone Scale}
\end{figure}

To focus on relevant skin areas, the method matches the dominant color in an image to the nearest class based on HSV values. This approach is chosen for its precision, speed, and efficiency in delivering accurate results. It operates by calculating the closest class using the CIEDE2000 formula, which quantifies the perceived differences between colors, enabling the system to identify the closest match based on the minimum distance.
The CIEDE2000 formula, which is designed for accurate color difference calculations, is as follows:

\[
\Delta E_{00} = \sqrt{\left( \frac{\Delta L'}{k_L S_L} \right)^2 + \left( \frac{\Delta C'}{k_C S_C} \right)^2 + \left( \frac{\Delta H'}{k_H S_H} \right)^2 + R_T \cdot \frac{\Delta C'}{k_C S_C} \cdot \frac{\Delta H'}{k_H S_H}}
\]

\subsection{Hair Color Classification}
\begin{figure}[h]
    \centering
\includegraphics[width=\textwidth]{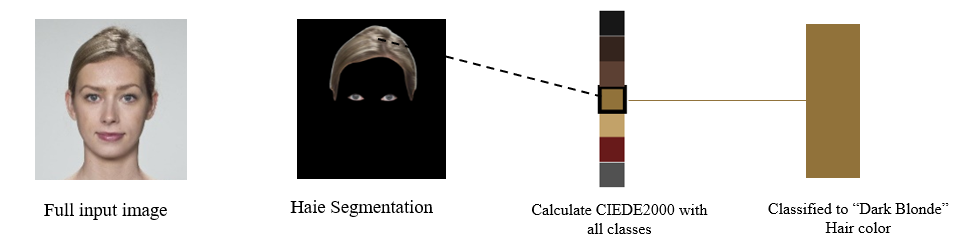} 
    \caption{ Hair Color Classification Processes Adapted from \cite{DeBruine2017}.}
    \label{fig:hair_method}
\end{figure}

The objective of this system is to classify the hair color by segmenting the hair region in an input image, extracting its color information, and comparing it to pre-defined categories of hair colors to find the closest match. Figure \ref{fig:hair_method}.

\newpage

\begin{enumerate}
    \item {\textbf{Input Image:} Accept an input image for hair color classification.}
    
    \item {\textbf{Hair Segmentation:} 
    Face detection by using RetinaFace and LaPa (Labeled Faces in the Wild with Parsing Annotations) for hair segmentation, then Hair mask creation, which is part of the model’s object detection output labels, to define boundaries around hair. \cite{yu2023faceperceiver}.}
    
    \item {\textbf{Hair Color Extraction:}  
    The system processes the segmented hair image by converting it into an array format for easy manipulation. Non-background pixels, representing the hair area, are retained, while irrelevant pixels are filtered out. The retained pixels are then gathered and prepared for color analysis over the hair region.}

    \item {\textbf{Dominant Color Calculation (K-means):} 
    Discover the most frequent color cluster in the hair area using K-means clustering (k = 3) for one color, and the hair distribution will be the dominant color that can be used.}

    \item {\textbf{Comparing Colors with Known Sets: } 
    First is color conversion to LAB for perceptual accuracy, converting the dominant and average RGB colors to LAB color space.
    Second, the LAB color distance calculation determines the distance using the CIEDE2000 formula between the predefined LAB color classes and the extracted LAB colors. Finally, aggregate distance is used to track the minimum distances for predefined categories and store the distances in dictionaries. The predefined color categories, similar to the skin tone scale method, developed through iterative artistic exploration and immediate adjustments, rather than following established color theory, pigment analysis, or expert advice, but instead through a creative process focused on attaining visually balanced compositions across varied styles.}

    \item {\textbf{Determining the Closest Color Category:} Calculate the combined distance metric by averaging the distances of the dominant and average colors for each category. Then identify the closest match by selecting the color category with the smallest total combined distance. }

\end{enumerate}


\subsection{Iris Color Classification}

The objective of this system is to classify the iris color by segmenting the iris region in an input image, extracting its color information, and comparing it to predefined categories of iris colors to find the closest match. The process involves the following steps:

\begin{figure}[H]  
    \centering
    \includegraphics[width=\linewidth]{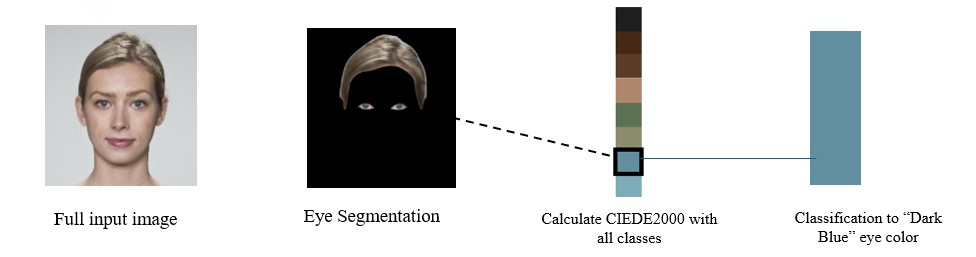}
    \caption{Iris Color Classification Processes Adapted from \cite{DeBruine2017}.}
    
    \label{fig:eye-method}
\end{figure}

\begin{enumerate}
    \item \textbf{Input Image}:  
    Accept an input image for iris color classification.
    
    \item \textbf{Iris Segmentation}:  
    The system uses Dlib’s face detection algorithm to identify facial regions in the input image. Following this, Dlib’s 68-point facial landmark predictor\cite{wu2019facial} is applied to accurately locate the eye regions. A circular mask is then generated around the iris to define its boundaries, effectively excluding irrelevant areas such as the pupil and sclera.
    
    \item \textbf{Iris Color Extraction}:  
    Convert the segmented iris region into an array for easier processing.  
    Retain only pixels that represent the iris, filtering out irrelevant parts such as the background, sclera, and eyelids.  
    Compute the average RGB color of the filtered pixels within the iris region.
    
    \item \textbf{Comparing Colors with Known Sets}:  
The computed average RGB color is converted to the CIE L*a*b* color space to ensure perceptual accuracy. The CIEDE2000 formula is then used to calculate the distance between the extracted LAB color and predefined LAB color categories. These calculated distances are aggregated to identify the closest match among the predefined categories. The custom scale used for these predefined categories was developed through a process of trial and error, without being based on any formal study, laboratory research, or professional calibration, but rather through practical experimentation aimed at optimizing model performance across diverse tones.

    \item \textbf{Determining the Closest Color Category}:  
    Select the closest match by identifying the predefined iris color category with the smallest Delta E value as the classified result.
\end{enumerate}

\subsection{Skin Undertone Classification}
This process classifies skin undertones as "Warm" or "Cool" using a wrist image. By leveraging image preprocessing, vein detection, LAB color conversion, and CIEDE2000 analysis, the system compares detected vein colors to predefined undertone values for precise classification.

\begin{figure}[h!]
    \centering
    \includegraphics[width=\linewidth]{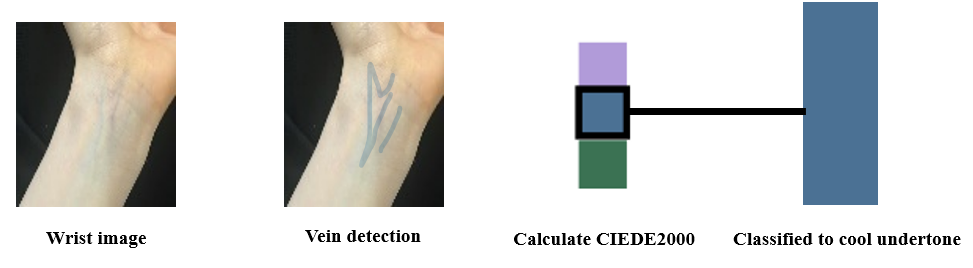}
    \caption{Skin Undertone Classification.}
    \label{fig:undertone}  
\end{figure}

\begin{enumerate}
    \item \textbf{Input Image:}  
    The system accepts an image of the wrist for skin undertone classification. It processes the image, performs color space conversions, and computes color differences to extract relevant features for classification.

    \item \textbf{Wrist Detection and Vein Analysis:}  
    The wrist image is transformed into the CIE L*a*b* color space to enhance color differentiation. The image is then separated into L, A, and B channels to focus on specific color attributes. Thresholds are applied in the LAB color space to identify skin and vein regions, creating separate masks for each. Bitwise operations are used to isolate the vein regions within the skin mask, enabling accurate analysis of the vein characteristics.

    \item \textbf{Noise Mitigation:}  
    Apply morphological operations (e.g., closing) to the vein mask to reduce noise, fill gaps, and improve the overall quality of the extracted region.

    \item \textbf{Calculation of Mean LAB Values:}  
    Compute the average LAB values for the isolated vein pixels to capture their key color characteristics.

    \item \textbf{Comparison with Predefined LAB Values and Undertone Classification:}  
The extracted LAB values from the hand veins are compared with predefined LAB references for warm ([70, 20, 40]) and cool ([60, -20, -30]) undertones using the CIEDE2000 formula to calculate the color difference. If the Delta E value is closer to the warm reference, the undertone is classified as “Warm” otherwise, it is classified as “Cool.”
These reference values were determined through trial and error, without relying on formal studies, and were selected to produce practical and usable results within the system.

    \item \textbf{Output Result:}  
    Display the final classification result, indicating whether the skin undertone is "Warm" or "Cool."
\end{enumerate}

\section{Evaluation}

\subsection{Evaluation Metrics}
To effectively evaluate the performance of our classifiers for Color Classification of Hair, Eyes, Skin Undertone, and Skin Overtone (Skin Tone), we utilized several key metrics that are vital for assessing reliability and accuracy. These metrics were particularly useful in the context of ensuring that our classifiers performed well across different categories and delivered meaningful results.

The core evaluation metrics used were derived from True Positives (TP), True Negatives (TN), False Positives (FP), and False Negatives (FN). True Positives indicate cases where the classifier correctly predicts a positive outcome, while True Negatives are instances where the classifier accurately predicts a negative outcome. False Positives (Type I errors) occur when the classifier incorrectly predicts a negative instance as positive, and False Negatives (Type II errors) arise when the classifier fails to identify a positive instance.

Based on these, the following metrics were employed to assess classifier performance:

\begin{itemize}
\item \textbf{Accuracy:} The ratio of correct predictions to the total number of predictions, provides an overall measure of the model's effectiveness.
    \[
    \text{Accuracy} = \frac{TP + TN}{TP + TN + FP + FN}
    \]
\item \textbf{Precision:} The proportion of true positive predictions out of all positive predictions made by the model, indicating the accuracy of positive classifications.
    \[
    \text{Precision} = \frac{TP}{TP + FP}
    \]

\item \textbf{Recall:} Also known as sensitivity, recall measures the ability of the model to correctly identify all positive instances.
    \[
    \text{Recall} = \frac{TP}{TP + FN}
    \]

\item \textbf{F1-Score:} A harmonic mean of precision and recall, providing a balanced measure when both metrics are important.
    \[
    \text{F1 Score} = 2 \cdot \frac{\text{Precision} \times \text{Recall}}{\text{Precision} + \text{Recall}}
    \]
\end{itemize}

\subsection{Experiment Design}
\subsubsection{Skin Tone Classification}

\begin{enumerate}
\item \textbf{Dataset}\\
    For this study, we utilized the dataset described in Section \ref{sec:Dataset/SkinToneDataset}, which comprises both a First Dataset and a Second Dataset. The \textbf{first dataset} is derived from three sources: \textbf{FFHQ} \cite{Karras2018}, \textbf{Face Research Lab London dataset} \cite{DeBruine2017}, and the \textbf{SFHQ} dataset \cite{SFHQDataset}. This composite dataset provided high-quality, diverse images categorized into 8 skin tone classes. The \textbf{second Dataset}, consisting of 400 images sourced exclusively from the FFHQ dataset, was utilized to address challenges such as lighting variations and shadows in the original datasets. By incorporating balanced contributions, this dataset further supported our analysis and experiments. To address challenges in consistent analysis, we experimented with various approaches to classify skin tones effectively. These challenges stemmed from variations in lighting, shadows, and other inconsistencies across the datasets. Specifically, we conducted experiments using \textbf{X-Means clustering} with different distance metrics and color spaces, both with and without applying a blurring effect to reduce noise in the images. The role of the datasets in the experiments was as shown in the Table \ref{tab:accuracy_blurring}:

    \begin{table}[h]
    \centering
    \begin{tabular}{lcc}
    \toprule
    \textbf{Method} & \textbf{Without Blurring} & \textbf{With Blurring} \\
    \midrule
    Euclidean Distance – RGB & 0.46 & 0.56 \\
    Delta E (CIE 2000) Distance – HSV & 0.42 & 0.73 \\
    Delta E (CIE 2000) Distance – RGB & 0.38 & 0.66 \\
    \bottomrule
    \end{tabular}
    \caption{Accuracy comparison with and without blurring effect for different distance methods.}
    \label{tab:accuracy_blurring}
    \end{table}

    From these results, it is evident that applying a blurring effect consistently improved classification accuracy across all distance metrics and color spaces, with the \textbf{CIE 2000 Distance – HSV} approach achieving the highest accuracy of \textbf{0.58}. To further enhance the results, we incorporated another dataset, which is the primary one for skin tone classification, thereby addressing some of the issues present in the secondary dataset. The addition of these datasets significantly improved the classification outcomes (detailed in the Results section \ref{sec:Results and discussion}). We also utilized the datasets for \textbf{Support Vector Machine (SVM)} classification. To ensure uniformity across classes, we applied data augmentation techniques, resulting in a total of \textbf{443 samples per class} and \textbf{3544 samples overall}. This balanced dataset provided a robust foundation for training and evaluation, contributing to more reliable classification performance.\\

\item \textbf{Classification Methods}\\

    Initially, we explored the CASCo library for skin tone analysis, focusing on mapping facial skin tones into the HSV color space \cite{rejon2023classification}. Non-skin regions such as eyes, hair, and teeth were filtered out using a Gaussian blur function. The remaining skin area was analyzed using k-means clustering to extract two dominant colors. These colors were categorized into a customizable class based on the closest color match, determined by the minimum weighted CIE 2000 distance \cite{sharma2005ciede2000}. Results were compiled in a results.csv file, including nine columns with detailed information, such as HEX values of the dominant colors in the facial area. However, the CASCo method proved less effective due to segmentation inaccuracies. Issues such as difficulties distinguishing skin from shadows, hair, and background colors affected the perceived accuracy of skin tones. Recognizing these limitations, we decided against using the PERLA color palette \cite{telles2013project}. While PERLA is valuable for specific applications, it was not designed to classify a diverse range of skin tones.
    
    To overcome these challenges, we refined our approach by incorporating advanced face detection and parsing techniques. This enhanced the isolation of skin tones from other elements in the image, such as shadows, hair, and background. The refinement improved classification accuracy by focusing solely on the skin color of detected faces, minimizing the impact of extraneous factors, and providing more reliable data for robust real-world applications. As a result, we utilized several classification methods, including: 
        \begin{itemize} 
        \item \textbf{X-Means with Euclidean Distance and RGB Color Space} 
        \item \textbf{X-Means with CIE 2000 Distance and HSV Color Space} 
        \item \textbf{X-Means with CIE 2000 Distance and RGB Color Space} 
        \end{itemize} 
    These methods shared a common configuration for the X-Means algorithm. The clustering process was initialized using K-means++ to identify initial cluster centers, and the Euclidean distance metric was used to evaluate cluster cohesion. The algorithm dynamically determined the optimal number of clusters based on the input data. Additionally, Gaussian blurring was applied to images to mitigate noise, with the kernel size proportional to the image dimensions and adjusted to ensure optimal smoothing. Gamma correction was also implemented to enhance image quality and normalize lighting inconsistencies.
    \\
    
    Additionally, we explored improving the classification process by trying a \textbf{Two-Stage Classification} approach. This method employed CIEDE2000 Distance and HSV with a blurring effect. The process started with the definition of main classes and their respective subclasses. The main classes were predefined with average color boundaries derived from our skin tone scale, where each main class was formed by combining two sequential classes from the 8-class skin tone scale. These averages represented the central tendencies of the main classes and served as reference points for determining the closest match during the first stage of classification. Once the main classes were established, subclasses within each main class were defined with distinct color values. These subclasses allowed for more precise differentiation within the identified main class in the second stage of classification. The two-stage process followed these steps:
        \begin{enumerate} 
        \item \textbf{First Stage: Determine the Main Class Boundary} \\ For each main class, the CIEDE2000 distance was calculated between the dominant color in the image and the average color of the main class. The main class with the smallest CIEDE2000 was selected as the broad classification.
        \item \textbf{Second Stage: Narrow Down to the Subclass} \\ Once the closest main class was identified, CIEDE2000 was calculated for each subclass within that main class. The subclass with the smallest CIEDE2000 was selected as the final classification.
        \end{enumerate} 
    Finally, we applied a \textbf{SVM} with features extracted using ResNet-18. The dataset was split into training (70\%), validation (15\%), and test (15\%) sets using a stratified approach. Hyperparameter tuning was performed using \textit{GridSearchCV} with a parameter grid of various \textbf{C} values ([0.1, 1, 10, 100]), \textbf{gamma} values ([1, 0.1, 0.01, 0.001]), and kernel types (\textit{rbf, poly, sigmoid}). The best parameters obtained were \textbf{C = 100}, \textbf{Gamma = 1}, and \textbf{Kernel = 'rbf'}. This method achieved a training accuracy of \textbf{0.82}, a validation accuracy of \textbf{0.76}, and a test accuracy of \textbf{0.75}. While effective, these results were slightly lower than those achieved using CIEDE2000 distance in the HSV color space with a blurring effect.
    
\end{enumerate}

\subsubsection{Hair Color Classification}
A variety of technical approaches were utilized to improve the classification of hair color images.
\begin{itemize}
    \item \textbf{First Approach:}This technique effectively classifies hair color in photos by using an organized methodology. The process begins with \textbf{Preprocessing Images}, where the input RGB images are converted to either Greyscale or HSV, depending on the color attributes required for analysis. This conversion enhances dataset augmentation and enables the system to focus on specific features, ultimately improving model performance. Following this, \textbf{Feature Extraction} is conducted using methods such as color histograms to identify the dominant colors in the hair region, extracting significant color features for detailed analysis and classification. Finally, \textbf{Classification Technique} involves training the k-Nearest Neighbors (k-NN) classifier using the extracted color data to categorize photos into predefined groups such as Black, Blonde, and Brown. This method is often selected for its simplicity and ease of interpretation, making it well-suited for preliminary studies. However, this approach has limitations: it shows \textbf{Low Resilience to Variability}, as the k-NN classifier struggles with changes in lighting, hair texture, or mixed colors, making it sensitive to noise and affecting prediction accuracy, especially with low-quality images. Moreover, there is a \textbf{Limited Number of Classes}, restricting the classification to a small set of predefined hair color groups, which may not capture the full diversity of hair colors.

    \item \textbf{Second Approach:}To better address the specific needs of our project, we modified the initial strategy in several ways. First, we incorporated \textbf{Hair Segmentation} using Facer to isolate the hair region and minimize background noise, thus enhancing the accuracy of feature extraction. Additionally, we employed \textbf{K-means Clustering in LAB Color Space} with $k=3$ instead of HSV, since the LAB space provides a perceptually uniform representation that better captures subtle variations in hair color. This change allows the model to notice finer differences across various shades. Furthermore, we introduced \textbf{ expanded color classes} to include a wider range of colors, making the model more adaptable and comprehensive for practical applications. Finally, instead of using a traditional classification model, we applied \textbf{Classification by Distance Calculation Using the CIEDE2000 Formula}, where the color distance between the clustered hair color and predefined color centers is computed. The category with the smallest color difference, according to the CIEDE2000 metric, is selected as the final predicted hair color.
\end{itemize}

\subsubsection{Iris Color Classification Approaches}
A combination of technical approaches was used to improve the classification of iris colors. The following subsections describe the various methodologies explored and their respective outcomes.

\begin{itemize}
\item \textbf{First Approach:} Improving Iris Color Classification Using Image Processing Techniques and Color Space
To enhance the iris color classification, a combination of image processing techniques was employed. The process began by calculating the distances between the iris and the pupil and using masks to isolate the iris region, aiming to reduce the influence of the pupil, which could appear in different colors (such as black or red) due to lighting conditions. While these steps helped improve feature extraction from the images, the significant variation in pupil size across different images affected the stability and precision of the extracted features.

To overcome these challenges, colors were transformed from the RGB color space to the CIE L*a*b* color space, which aligns better with human color perception. This transformation allowed for the classification of iris colors into three main categories: dark, medium, and light, further divided into nine subcategories. Although this approach showed a slight improvement in overall performance, the overlap of color values between the medium and light categories posed an obstacle to precise classification. This emphasizes the need for more advanced techniques to address these challenges and achieve more accurate results.

\item\textbf{Second Approach:} Enhancing Iris Color Detection with Gaussian Blur and Self-Attention
Gaussian Blur (GBlur) was applied to the iris region to minimize noise and smooth out fine details, facilitating the extraction of the dominant color while using masks to exclude the pupil and focus solely on the iris. This approach successfully identified the primary iris color but faced challenges with multi-colored irises, such as hazel, which often feature intricate gradients. Additionally, a self-attention mechanism was employed to enhance classification by concentrating on the most relevant iris features during feature extraction. While this method improved accuracy overall, it occasionally misclassified iris colors when relying on a limited pixel sample that did not fully capture the entire iris's color spectrum. These iterative experiments highlighted both the strengths and limitations of each method, contributing to the refinement of techniques for more precise iris color detection.

\item\textbf{Third Approach:} Dlib's Pre-Trained Face Detector and 68-Point Facial Landmark Predictor for Iris Color Detection
This approach incorporates several advanced features to ensure efficiency and accuracy in detecting and classifying iris colors. It utilizes Dlib's pre-trained face detector and 68-point facial landmark predictor to quickly and accurately identify facial regions and locate key features like the eyes. The iris region is precisely isolated using geometric calculations based on eye landmarks, which determine the center and radius, minimizing interference from surrounding areas. A circular mask is applied to isolate the iris, ensuring that only relevant regions are analyzed. The extracted iris color, initially in RGB format, is transformed into the CIE L*a*b* color space for better perceptual accuracy. To classify the iris color, the CIEDE2000 formula is used to measure the perceptual difference between the detected color and predefined color standards, ensuring precise matching. Additionally, the system processes both eyes separately and calculates an average color for a unified classification.
\end{itemize}
\subsubsection{Skin Undertone Classification}

The classification of skin undertones into "Warm" or "Cool" is achieved using image processing techniques and LAB color space analysis. Two approaches were employed to refine the classification process, each with distinct methodologies and outcomes :
\begin{itemize} 

\item\textbf{First Approach:} This method focuses on classifying skin undertones as "Warm" or "Cool" by applying preprocessing techniques, LAB transformations, and cosine similarity for analysis. The process begins with resizing images for computational efficiency and applying Gaussian blur to reduce noise. Adaptive thresholding is then employed to emphasize vein patterns, while morphological operations improve vein connectivity and eliminate background artifacts, ensuring clearer vein structures. The LAB color space is used to enhance color differentiation, with thresholds defined to detect skin and vein regions. Veins are isolated by subtracting the skin mask from the vein mask. The mean LAB values of the isolated vein regions are calculated to represent their color properties. Cosine similarity is then used to compare these detected values with predefined LAB values for warm and cool undertones. Classification is determined based on the highest cosine similarity score, categorizing undertones as "Warm" (green/olive hues) or "Cool" (blue/purple hues). Despite its efficiency, the approach encountered challenges, particularly in misclassifying veins with colors near green or blue, leading to ambiguity between "Warm" and "Cool" classifications. These limitations highlighted the need for refining vein isolation techniques and improving the accuracy of similarity calculations for enhanced performance.

\item\textbf{Second Approach:} Building on the first approach, this method incorporates the CIEDE2000 formula for more precise undertone classification. The process begins with resizing images and converting them to the LAB color space to improve color differentiation and align with human perception. LAB thresholds are used to detect skin and vein regions, creating skin and vein masks, followed by bitwise operations to isolate vein regions for focused analysis. Morphological operations enhance vein quality by reducing noise, filling gaps, and ensuring clearer vein connectivity. The key advancement in this approach lies in the use of the CIEDE2000 formula, which measures perceptual color differences. The mean LAB values of the isolated vein regions are compared to predefined LAB references for warm undertones ([70, 20, 40]) and cool undertones ([60, -20, -30]). The classification is determined based on the smallest Delta E value, categorizing undertones as "Warm" or "Cool." This approach addresses ambiguities in the first method, enhancing accuracy by using the CIEDE2000 formula to focus on perceptual differences. Sharing preprocessing steps with the first approach improves robustness and reduces misclassification for more reliable results.
\end{itemize}

\section{Results and Discussion}
\label{sec:Results and discussion}

\subsection{Skin Tone Classification}

The three approaches were evaluated to improve the accuracy of skin tone classification, using two datasets comprising primary and secondary images. Each approach was systematically assessed to determine its effectiveness. The results without blurring showed suboptimal performance, whereas incorporating blurring significantly improved the classification results. Blurring can be highly beneficial for extracting skin tone as it reduces the influence of fine details and noise, enabling a focus on the overall color distribution of the skin. This technique plays a crucial role in improving the accuracy and consistency of skin tone classification. Among these approaches, the best results were achieved by combining blurring with the X-Means clustering algorithm, the CIE 2000 distance metric, and the HSV color space, attaining the highest accuracy of 0.80. This enables the classifier to effectively distinguish between the eight classes of skin tone classification.

\begin{table}[h]
\centering
\resizebox{\textwidth}{!}{ 
\begin{tabular}{@{}llcccccccc@{}}
\toprule
\multirow{2}{*}{\textbf{Method}} &
\multirow{2}{*}{\textbf{Metric}} &
\multicolumn{2}{c}{\textbf{Primary}} &
\multicolumn{2}{c}{\textbf{Secondary}} \\ \cmidrule(lr){3-4} \cmidrule(lr){5-6}
& & \textbf{Without Blur} & \textbf{With Blur} & \textbf{Without Blur} & \textbf{With Blur} \\ \midrule
\multirow{4}{*}{X-Means with Euclidean Distance and RGB Color Space}
& Accuracy  & 0.46 & 0.56 & 0.49 & 0.53 \\
& Precision & 0.40 & 0.56 & 0.45 & 0.54 \\
& Recall    & 0.46 & 0.56 & 0.49 & 0.53 \\
& F1 Score  & 0.39 & 0.55 & 0.43 & 0.52 \\ \midrule

\multirow{4}{*}{\textbf{X-Means with CIE 2000 Distance and HSV Color Space}}
& Accuracy  & 0.42 & 0.73 & 0.40 & \textbf{0.80} \\
& Precision & 0.26 & 0.76 & 0.24 & \textbf{0.81} \\
& Recall    & 0.42 & 0.73 & 0.40 & \textbf{0.80}\\
& F1 Score  & 0.31 & 0.73 & 0.29 & \textbf{0.80} \\\midrule

\multirow{4}{*}{X-Means with CIE 2000 Distance and RGB Color Space}
& Accuracy  & 0.38 & 0.66 & 0.35 & 0.65 \\
& Precision & 0.22 & 0.67 & 0.20 & 0.68 \\
& Recall    & 0.38 & 0.66 & 0.35 & 0.65 \\
& F1 Score  & 0.27 & 0.65 & 0.24 & 0.64 \\ \bottomrule
\end{tabular}
}
\caption{Performance Metrics for X-Means Methods}
\end{table}

\begin{table}[h]
\centering
\begin{minipage}{0.6\textwidth}
\centering
\resizebox{\textwidth}{!}{%
\begin{tabular}{@{}lcc@{}}
\toprule
\textbf{Method}                     & \textbf{Primary}                & \textbf{Secondary}                \\ \midrule
\multirow{4}{*}{Two-Stage Classification} 
& Accuracy: 0.64 & Accuracy: 0.64 \\
& Precision: 0.69 & Precision: 0.69 \\
& Recall: 0.64    & Recall: 0.64    \\
& F1 Score: 0.64  & F1 Score: 0.64  \\ \bottomrule
\end{tabular}
}
\caption{Performance Metrics for Two-Stage Classification}
\end{minipage}%
\hfill
\begin{minipage}{0.35\textwidth}
\centering
\resizebox{\textwidth}{!}{%
\begin{tabular}{ll}
\toprule
\textbf{SVM with ResNet18} & \textbf{Primary} \\ \midrule
Accuracy: & 0.76 \\
Precision: & 0.76 \\
Recall: & 0.76 \\
F1 Score: & 0.76 \\ \bottomrule
\end{tabular}%
}
\caption{Performance Metrics for SVM with ResNet18}
\end{minipage}
\end{table}

\subsection{Color Classification For the Hair, Eyes, and Undertone}
To evaluate the classification accuracy, a test set was created by manually collecting 10 images per color class. A truth table was constructed for each image to record the classification result against the ground truth. The performance was measured as follows:

\begin{enumerate}
    \item \textbf{Hair Color classification}\\
    The classification accuracies in Table \ref{tab:classification_results} can be analyzed based on multiple potential factors. As a result of their unique characteristics, categories such as dark brown, black, and gray achieved high accuracy. On the other hand, blonde, brown, and dark blonde showed low accuracy, likely due to underrepresentation and overlapping characteristics, making them more difficult to distinguish, especially across comparable tones such as brown and dark blonde. “Light colors such as blonde and dark blonde are also more difficult to classify due to their wider range of lightness and shading, while dark colors such as black and dark brown are simpler because they have less contrast in intensity and hue.
    
    \begin{table}[H]
        \centering
        \begin{tabular}{lc}
            \toprule
            \textbf{Color Class} & \textbf{Accuracy (\%)} \\
            \midrule
            Black & 80\% \\
            Blonde & 50\% \\
            Brown & 20\% \\
            Dark Blonde & 30\% \\
            Dark Brown & 90\% \\
            Grey & 50\% \\
            Red & 70\% \\
            \bottomrule
        \end{tabular}
        \caption{Hair Color Classification Results}
        \label{tab:classification_results}
    \end{table}

    \item\textbf{Eye Color Classification}\\
    Table \ref{tab:eye_classification_results}'s classification accuracies provide light on the eye color classification. The uniqueness of each eye color, possible overlaps between related hues, and the generalisability of the model are some of the criteria that can be used to evaluate these findings. The following summarises the potential causes of the observed results:
    Black, Dark Blue, and Dark Hazel (Brown) all had perfect accuracy (100\%) because of their unique characteristics and little overlap with other classes. Though there may be sporadic misclassifications due to overlapping shades or reflection artifacts, eye colors such as Light Blue, Dark Green, Light Green, and Light Hazel (Brown) demonstrated great accuracy (80\%–90\%). Light Blue may be mistaken for similar tones like Dark Blue. Grey, which had the lowest accuracy (70\%), probably had trouble differentiating itself because it looked like Light Blue or Light Green.
    \\

    \begin{table}[h!]
        \centering
        \begin{tabular}{lcc}
            \toprule
            \textbf{Eye Color} & \textbf{Accuracy (\%)} \\
            \midrule
            Dark Blue & 100\% \\
            Light Blue & 80\% \\
            Dark Green & 90\% \\
            Light Green & 90\% \\
            Dark Hazel(Brown) & 100\% \\
            Light Hazel (Brown) & 90\% \\
            Black & 100\% \\
            Gray & 70\% \\
    
            \bottomrule
        \end{tabular}
        \caption{Eye Color Classification Results}
        \label{tab:eye_classification_results}
        \end{table}

    \item\textbf{Undertone Color classification}\\
    According to the undertone classification table \ref{tab:undertone_results}, the accuracy for Warm undertones is 80\%, while the accuracy for Cool undertones is 70\%. Showing generally positive and almost identical results.
    
    \begin{table}[h!]
        \centering
        \begin{tabular}{lcc}
            \toprule
            \textbf{Undertone} & \textbf{Accuracy (\%)} \\
            \midrule
            Cool & 70\% \\
            Warm & 80\% \\
            \bottomrule
        \end{tabular}
        \caption{Undertone Classification Results}
        \label{tab:undertone_results}
    \end{table}
\end{enumerate}

\section{Conclusion}
This study introduces a comprehensive and intelligent framework for the classification of key human attributes—skin tone, hair color, iris color, and undertone—using advanced imaging and machine learning techniques. By integrating multi-stage image processing, color space transformations, and perceptual color difference metrics, the system achieves high accuracy in recognizing subtle human color variations, even under diverse lighting conditions. The use of both face and wrist images enables a more holistic and inclusive analysis, accommodating the complexity of natural human features. The promising performance, with classification accuracies reaching up to 80\% using the Delta E–HSV method with Gaussian blur, demonstrates the effectiveness of our approach. Importantly, we did not rely on a single color space across all classifications. Instead, we used both HSV and LAB depending on the nature of the attribute. Each feature—such as skin, hair, eyes, and undertone—requires a specific representation for optimal results. These choices were driven by practical experimentation and what yielded the best accuracy in each case. Moreover, the results and techniques were made accessible to facilitate researchers’ access and encourage continued research and development in this field. Overall, this work contributes a valuable tool for personalized color-based analysis, with potential applications in beauty technology, digital identity, and human-centered design.

\bibliographystyle{unsrt}  
\bibliography{references}

\begin{thebibliography}{10}

\bibitem{bunte2011learning}
Kerstin Bunte, Michael Biehl, Marcel~F Jonkman, and Nicolai Petkov.
\newblock Learning effective color features for content based image retrieval in dermatology.
\newblock {\em Pattern Recognition}, 44(9):1892--1902, 2011.

\bibitem{campbell2020picture}
Mary~E Campbell, Verna~M Keith, Vanessa Gonlin, and Adrienne~R Carter-Sowell.
\newblock Is a picture worth a thousand words? an experiment comparing observer-based skin tone measures.
\newblock {\em Race and Social Problems}, 12:266--278, 2020.

\bibitem{chen2016skin}
Wei Chen, Ke~Wang, Haifeng Jiang, and Ming Li.
\newblock Skin color modeling for face detection and segmentation: a review and a new approach.
\newblock {\em Multimedia Tools and Applications}, 75:839--862, 2016.

\bibitem{dash2020biometric}
Rosalin Dash.
\newblock Biometric face skintone data augmentation using a generative adversarial network.
\newblock 2020.

\bibitem{DeBruine2017}
Lisa DeBruine and Benedict Jones.
\newblock Face research lab london set.
\newblock \url{https://doi.org/10.6084/m9.figshare.5047666.v5}, 2017.
\newblock Dataset.

\bibitem{deng2020retinaface}
Jiankang Deng, Jia Guo, Evangelos Ververas, Irene Kotsia, and Stefanos Zafeiriou.
\newblock Retinaface: Single-shot multi-level face localisation in the wild.
\newblock In {\em Proceedings of the IEEE/CVF Conference on Computer Vision and Pattern Recognition}, pages 5203--5212, 2020.

\bibitem{gulati2023beautifai}
Kshitij Gulati, Gaurav Verma, Mukesh Mohania, and Ashish Kundu.
\newblock Beautifai-personalised occasion-based makeup recommendation.
\newblock In {\em Asian Conference on Machine Learning}, pages 407--419. PMLR, 2023.

\bibitem{hassan2023hsv}
Enas~Kh Hassan and Jamila~Harbi Saud.
\newblock Hsv color model and logical filter for human skin detection.
\newblock In {\em AIP Conference Proceedings}, volume 2457. AIP Publishing, 2023.

\bibitem{heldreth2023skin}
Courtney~M Heldreth, Ellis~P Monk, Alan~T Clark, Candice Schumann, Xango Eyee, and Susanna Ricco.
\newblock Which skin tone measures are the most inclusive? an investigation of skin tone measures for artificial intelligence.
\newblock {\em ACM Journal on Responsible Computing}, 2023.

\bibitem{ibraheem2012understanding}
Noor~A Ibraheem, Mokhtar~M Hasan, Rafiqul~Z Khan, and Pramod~K Mishra.
\newblock Understanding color models: a review.
\newblock {\em ARPN Journal of science and technology}, 2(3):265--275, 2012.

\bibitem{ISLAM2024104046}
ABM~Rezbaul Islam, Ali Alammari, and Bill Buckles.
\newblock Human skin detection: An unsupervised machine learning way.
\newblock {\em Journal of Visual Communication and Image Representation}, 98:104046, 2024.

\bibitem{Jackson1980}
Carole Jackson.
\newblock {\em Color Me Beautiful}.
\newblock Ballantine Books, New York, 1980.

\bibitem{srt70088}
Geunho Jung, Semin Kim, and Sangwook Yoo.
\newblock Skin tone analysis through skin tone map generation with optical approach and deep learning.
\newblock {\em Skin Research and Technology}, 30(10):e70088, 2024.

\bibitem{Karras2018}
Tero Karras, Samuli Laine, and Timo Aila.
\newblock A style-based generator architecture for generative adversarial networks.
\newblock {\em arXiv preprint arXiv:1812.04948}, 2018.
\newblock \url{https://arxiv.org/abs/1812.04948}.

\bibitem{article2}
Shristi Khanal, Prasad P.W.C, Abeer Alsadoon, and Angelika Maag.
\newblock A systematic review: machine learning based recommendation systems for e-learning.
\newblock {\em Education and Information Technologies}, 25:1--30, 07 2020.

\bibitem{KIM2023108247}
Boram Kim, Juhyun Lee, Sungmi Park, and Hyeon-Jeong Suk.
\newblock Method and analysis of color changes of facial skin after applying skin makeup.
\newblock {\em Vision Research}, 209:108247, 2023.

\bibitem{kolkur2017human}
Seema Kolkur, Dhananjay Kalbande, P~Shimpi, Chaitanya Bapat, and Janvi Jatakia.
\newblock Human skin detection using rgb, hsv and ycbcr color models.
\newblock {\em arXiv preprint arXiv:1708.02694}, 2017.

\bibitem{mamat2018silhouette}
Abd~Rasid Mamat, Fatma~Susilawati Mohamed, Mohamad~Afendee Mohamed, Norkhairani~Mohd Rawi, and Mohd~Isa Awang.
\newblock Silhouette index for determining optimal k-means clustering on images in different color models.
\newblock {\em Int. J. Eng. Technol}, 7(2):105--109, 2018.

\bibitem{nidhal2009skin}
KA~Nidhal, SD~Nizar, and AA~Zaid.
\newblock Skin texture recognition using neural networks.
\newblock In {\em Conference: ACIT}, volume~9, 2009.

\bibitem{inbook}
Sandeep Raghuwanshi and R.~Pateriya.
\newblock {\em Recommendation Systems: Techniques, Challenges, Application, and Evaluation: SocProS 2017, Volume 2}, pages 151--164.
\newblock 01 2019.

\bibitem{rejon2023classification}
Ren{\'e}~Alejandro Rej{\'o}n~Pi{\~n}a and Chenglong Ma.
\newblock Classification algorithm for skin color (casco): A new tool to measure skin color in social science research.
\newblock {\em Social Science Quarterly}, 104(2):168--179, 2023.

\bibitem{SFHQDataset}
SelfishGene.
\newblock Sfhq dataset.
\newblock \url{https://github.com/SelfishGene/SFHQ-dataset}, 2024.
\newblock GitHub Repository.

\bibitem{sharma2005ciede2000}
Gaurav Sharma, Wencheng Wu, and Edul~N Dalal.
\newblock The ciede2000 color-difference formula: Implementation notes, supplementary test data, and mathematical observations.
\newblock {\em Color Research \& Application: Endorsed by Inter-Society Color Council, The Colour Group (Great Britain), Canadian Society for Color, Color Science Association of Japan, Dutch Society for the Study of Color, The Swedish Colour Centre Foundation, Colour Society of Australia, Centre Fran{\c{c}}ais de la Couleur}, 30(1):21--30, 2005.

\bibitem{sobhan2022subject}
Masrur Sobhan, Daniela Leizaola, Anuradha Godavarty, and Ananda~Mohan Mondal.
\newblock Subject skin tone classification with implications in wound imaging using deep learning.
\newblock In {\em 2022 International Conference on Computational Science and Computational Intelligence (CSCI)}, pages 1640--1645. IEEE, 2022.

\bibitem{telles2014pigmentocracies}
Edward Telles.
\newblock {\em Pigmentocracies: Ethnicity, race, and color in Latin America}.
\newblock UNC Press Books, 2014.

\bibitem{telles2013project}
Edward Telles et~al.
\newblock Project on ethnicity and race in latin america (perla).
\newblock {\em Pigmentocracies: Ethnicity, Race and Color in Latin America}, 2013.

\bibitem{ware2020racial}
Olivia~R Ware, Jessica~E Dawson, Michi~M Shinohara, and Susan~C Taylor.
\newblock Racial limitations of fitzpatrick skin type.
\newblock {\em Cutis}, 105(2):77--80, 2020.

\bibitem{Weiser}
Mark Weiser and John~Seely Brown.
\newblock {\em The coming age of calm technolgy}, pages 75--85.
\newblock Copernicus, New York, NY, USA, 1997.

\bibitem{wu2019facial}
Yue Wu and Qiang Ji.
\newblock Facial landmark detection: A literature survey.
\newblock {\em International Journal of Computer Vision}, 127(2):115--142, 2019.

\bibitem{yu2023faceperceiver}
Jiachen Yu, Jiankang Xu, Shiyang Yan, Yibo Feng, Ran Cai, Li~Zhang, Ying Shan, Xiaoguang Chen, and Meiguang Kan.
\newblock Faceperceiver: Face parsing in a transformer with decomposed attention.
\newblock In {\em Proceedings of the IEEE/CVF Conference on Computer Vision and Pattern Recognition}, pages 18487--18497, 2023.

\bibitem{zheng2022general}
Yinglin Zheng, Hao Yang, Ting Zhang, Jianmin Bao, Dongdong Chen, Yangyu Huang, Lu~Yuan, Dong Chen, Ming Zeng, and Fang Wen.
\newblock General facial representation learning in a visual-linguistic manner.
\newblock In {\em Proceedings of the IEEE/CVF conference on computer vision and pattern recognition}, pages 18697--18709, 2022.

\bibitem{Zhu2022}
Wanshan Zhu, Peng Sang, and Yifan He.
\newblock Facial skin colour classification using machine learning and hyperspectral imaging data.
\newblock {\em IET Image Processing}, 16:509--520, 2 2022.

\bibitem{Zuo2017}
Haiqiang Zuo, Heng Fan, Erik Blasch, and Haibin Ling.
\newblock Combining convolutional and recurrent neural networks for human skin detection.
\newblock {\em IEEE Signal Processing Letters}, 24:289--293, 3 2017.

\end{thebibliography}

\end{document}